\documentclass[10pt,twocolumn,letterpaper]{article}

\usepackage{cvpr}              
\usepackage{graphicx}
\usepackage{amsmath}
\usepackage{amssymb}
\usepackage{booktabs}

\usepackage[pagebackref,breaklinks,colorlinks]{hyperref}

\usepackage[capitalize]{cleveref}
\crefname{section}{Sec.}{Secs.}
\Crefname{section}{Section}{Sections}
\Crefname{table}{Table}{Tables}
\crefname{table}{Tab.}{Tabs.}

\def\cvprPaperID{29} 
\def\confName{CVPR}
\def\confYear{2023}

\begin{document}

\title{Context-Preserving Two-Stage Video Domain Translation \\ 
        for Portrait Stylization}

\author{Doyeon Kim$^1$, Eunji Ko$^1$, Hyunsu Kim$^2$, Yunji Kim$^2$, Junho Kim$^2$, Dongchan Min$^1$, \\Junmo Kim$^1$, Sung Ju Hwang$^1$\\
Korea Advanced Institute of Science and Technology (KAIST)$^1$ \\
NAVER AI Lab$^2$ \\
{\tt\small \{doyeon\_kim, kosu7071, alsehdcks95, junmo.kim, sjhwang82\}@kaist.ac.kr}\\
{\tt\small \{hyunsu1125.kim, yunji.kim, jhkim.ai\}@navercorp.com}
}

\twocolumn[{%
\renewcommand\twocolumn[1][]{#1}%
\maketitle
\begin{center}
    \centering
    \captionsetup{type=figure}
    \includegraphics[width=0.98\textwidth]{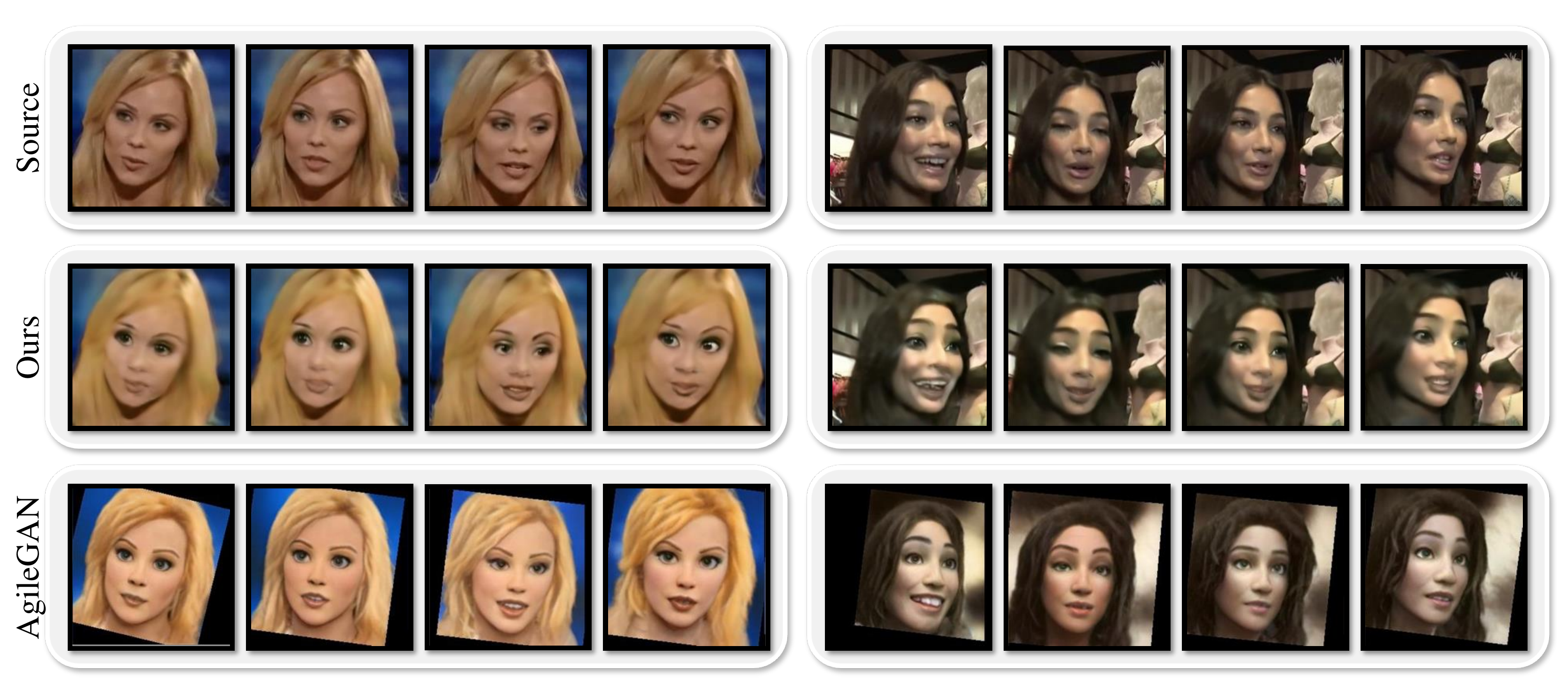}
    \captionof{figure}{Stylized video frames from ours and prior approach~\cite{song2021agilegan}. In this work, we tackle the video stylization problem while focusing on the information loss (e.g. identity, eye gaze direction, hair movement, ...) and temporal inconsistency limitation of the existing StyleGAN-based portrait stylization method.} 
\end{center}%
}]

\begin{abstract}

Portrait stylization, which translates a real human face image into an artistically stylized image, has attracted considerable interest and many prior works have shown impressive quality in recent years. 
However, despite their remarkable performances in the image-level translation tasks, prior methods show unsatisfactory results when they are applied to the video domain.
To address the issue, we propose a novel two-stage video translation framework with an objective function which enforces a model to generate a temporally coherent stylized video while preserving context in the source video.
Furthermore, our model runs in real-time with the latency of 0.011 seconds per frame and requires only 5.6M parameters, and thus is widely applicable to practical real-world applications.

\end{abstract}

\section{Introduction}
\label{sec:intro}

\emph{Portrait stylization}, which aims to transform a real human face image into one with an artistic style, is widely used in various fields such as advertising, animation production, or filmmaking.
Yet, this process requires a lot of effort and time even for a skilled artist, which led to the introduction of automatic portrait stylization~\cite{kim2019u, nizan2020breaking, chong2021gans, li2021anigan} methods based on deep neural networks that can obtain plausible results without human intervention. 
Especially, StyleGAN~\cite{karras2019style}, which is a generative adversarial network (GAN) based on style latents, has greatly improved the performance of human face generation and extended the range of applications as it allows to modify the style of the generated images with a simple modification of style vectors which modulate the generator.
To name a few, \cite{yang2022pastiche, chong2021jojogan, song2021agilegan} have shown impressive results in \emph{image-level} translation, in which they utilize GAN inversion techniques~\cite{alaluf2021restyle, richardson2021encoding,kingma2013auto} to map the source image into latent space and decode it with a StyleGAN to generate a stylized image. 
However, despite their impressive performances in \emph{image-level} portrait stylization, they show limited capability in the video-level translation as can be seen in Fig.~\ref{sec:intro}, where the result of the baseline is generated by frame-by-frame translation.

The reasons for such suboptimal performances are as follows: First, the majority of previous image-level methods require geometric constraints such as facial landmark alignment to perform the translation, which leads to the generation of unnatural videos since the landmarks of the human face should be in a fixed position.
Second, the loss of critical information occurs in the GAN inversion process.
StyleGAN-based methods heavily rely on GAN inversion techniques to encode source 
images into the latent feature space, but they might yield images that do not preserve the original content of the source image.
For example, previous methods often generate translations with different movements of facial features such as mouth and eyes or the pose of the face from a source video, which are the important properties that the video transfer should preserve. 
Furthermore, the identity loss frequently occurs due to incomplete inversion.
Finally, there is no consideration for temporal dimension in existing methods based on the frame-by-frame translation. Image-level translation generates consecutive output frames by observing the current source frame only.
Thus, it is challenging to produce a smooth temporal motion and detect temporal noise.


In this work, we propose a novel method to create a stylized video with a real human face video as an input which addresses the aforementioned limitations of previous StyleGAN-based approaches.
We suggest a two-stage training scheme to decompose the problem into two manageable sub-problems, domain transfer and video generation.
In the first stage, we train the mapping network which not only learns to map the source domain to the target domain but also effectively delivers contextual information from input data to enhance spatial correlation and avoid the usage of GAN inversion. 
For the second stage, we naturally expand the image generator to the video domain by designing a sequential refiner which incorporates multiple consecutive frames of the network. 
Our sequential refiner enables the whole network to consider given intermediate output frames from the image generator and produce temporally coherent output frames with the suggested objective function.
Moreover, our model is suitable for practical real-world applications as it requires only a small number of parameters and runs in real-time speed.

\begin{figure}[t]
    \centering
    \begin{tabular}{ccc}
    \hspace{-0.4cm}
    \begin{minipage}{0.24\textwidth}\includegraphics[trim=0cm 0cm 0cm 0cm,clip,width=\linewidth]{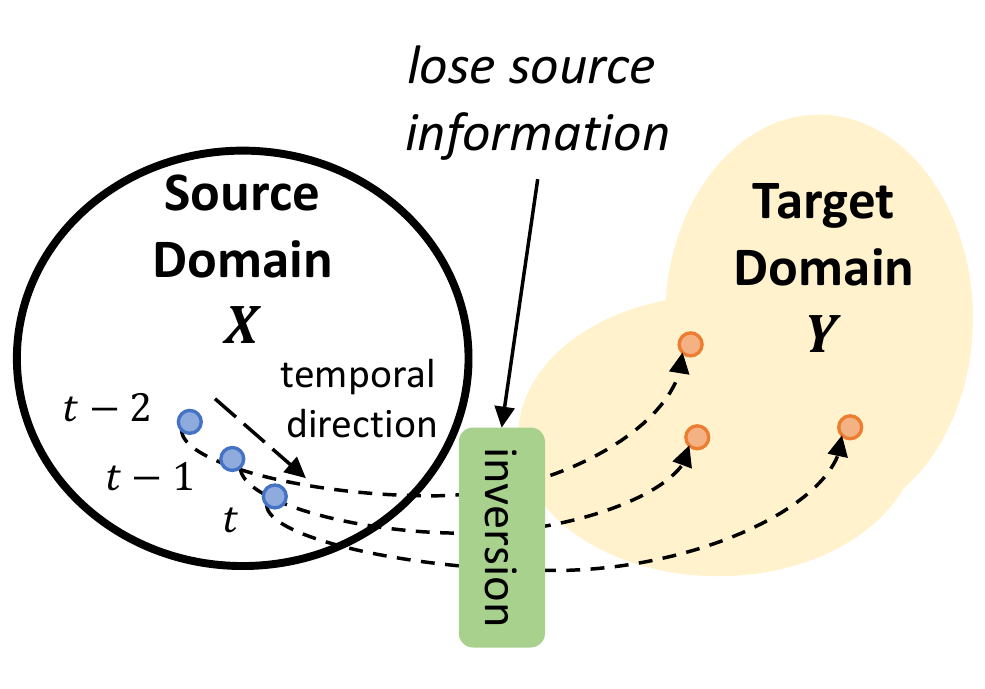}\end{minipage} &
    \hspace{-0.4cm}
    \begin{minipage}{0.24\textwidth}\includegraphics[trim=0cm 0cm 0cm 0cm,clip,width=\linewidth]{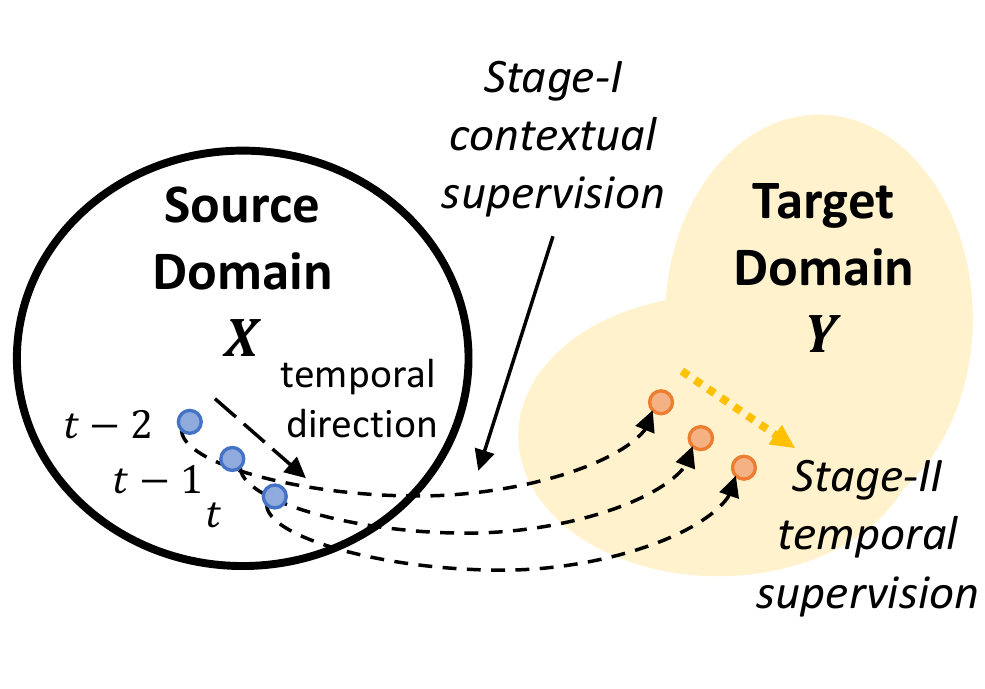}\end{minipage}
    \\
    \vspace{0.5cm}
    (a) Previous portrait stylization & 
    (b) Ours \\
    \vspace{-1cm}
\end{tabular}
\caption{Comparison between previous image-level StyleGAN-based portrait stylization and our portrait video style transfer. }
    \vspace{-0.5cm}
\label{fig:concept}
\end{figure}

\section{Methods}

Let $\boldsymbol{x}=\{x_1, x_2, \dots, x_T\}$ be an input portrait video with $T$ frames where $\boldsymbol{x}$ follows the distribution of source domain $X$.
The purpose of this work is to train a network $G$ to translate the input video $\boldsymbol{x}$ to $\hat{\boldsymbol{y}}=G(\boldsymbol{x})$, where $\hat{\boldsymbol{y}}=\{\hat{y_1}, \hat{y_2}, \dots, \hat{y_T}\}$
follows the distribution of target domain $Y$.
Specifically, we aim to train a model that 1) preserves the identity and facial attributes of source video $\boldsymbol{x}$
and 2) enforces temporal consistency in the output video $\hat{\boldsymbol{y}}$.

\noindent\textbf{Stage-I: Context-Preserving Domain Translation.}
In the first stage, we train the network to translate an image $x$ from the source domain $X$ to a stylized image $\hat{y}$ that follows the distribution of target domain $Y$.
We observe that StyleGAN-based methods work fairly well for the image-level portrait stylization task, but a significant amount of contextual information is lost as illustrated in Fig.~\ref{fig:concept} due to GAN inversion processes that map the source images into the latent space.
Hence, to minimize this loss, we employ the network $G_{X \rightarrow Y}$ as a domain translator which does not require the mapping of an image to the latent space to retain contextual information as much as possible.
Specifically, we train two networks, StyleGAN-based network $G_{Y}$ that learns to generate images of the target domain $Y$ from the normal distribution $\mathcal{N}(0,I)$, and U-Net-based~\cite{unet} network $G_{X \rightarrow Y}$ that translates an input image to be close to the output of $G_{Y}$.
We first train a StyleGAN-based generator $G_Y$ that follows the distribution of the target domain $Y$ by fine-tuning the pre-trained StyleGAN $G_X$ that generates images of the source domain $X$.
Since $G_X$ and $G_Y$ can generate images that are contextually identical when they are fed with the same random value $z$ as an input, we can construct the dataset of a pair of real and stylized images {$(\hat{x}, \hat{y})$} and use as a form pseudo supervision.

\begin{figure*}[t]
\centering
\includegraphics[width=1\linewidth]{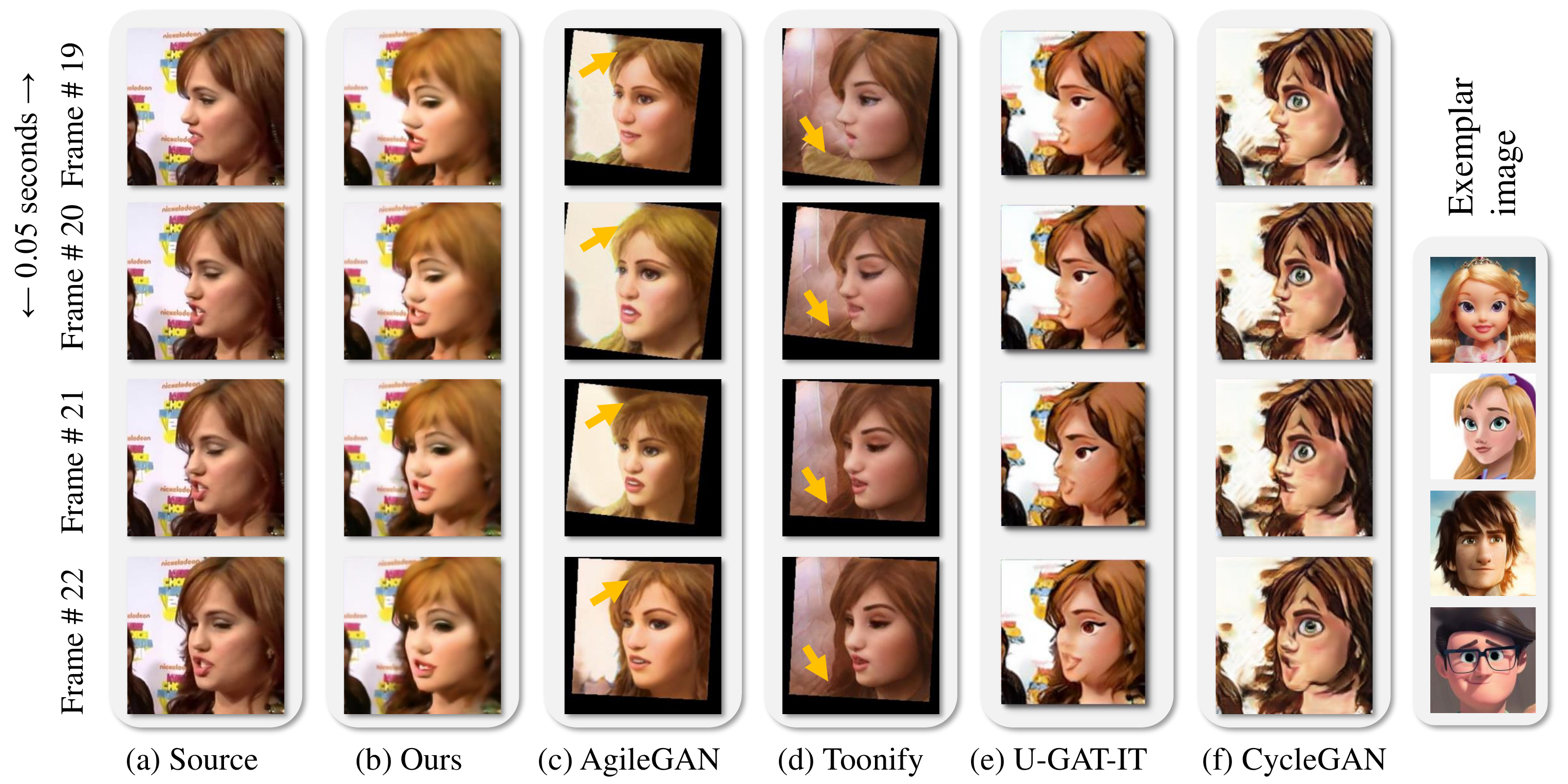}
\vspace{-0.4cm}
   \caption{Qualitative comparison results with previous works. Source frames are captured with 20 fps which means that the time interval between each frame is 0.05 seconds. 
    } 
\label{fig:comparison}
\end{figure*}

Then, we train a mapping network $G_{X \rightarrow Y}$ with previously constructed dataset $(\hat{x}, \hat{y})$ to perform domain translation. 
First, we apply adversarial loss $\mathcal{L}_{adv}$ with $\hat{x}$ and $\hat{y}$ to learn to translate in the whole domain level $X \rightarrow Y$. 
Furthermore, we calibrate the individual data point using $\mathcal{L}_{content}$ to reduce the heavily translated area and adjust the image so that it does not deviate drastically from the source image. We utilize two terms, reconstruction loss and perceptual loss. Reconstruction loss $\mathcal{L}_{recon}$ is formulated as the L1 distance between $G_{X \rightarrow Y}(\hat{x})$ and $\hat{y}$ in the image level space, and perceptual loss $\mathcal{L}_{perc}$ is calculated following \cite{mechrez2018contextual}.

\noindent\textbf{Stage-II: Sequential Video Frame Generation.} In this stage, we aim to maintain the temporal consistency between a source video $\boldsymbol{x}$ and a translated video $\hat{\boldsymbol{y}}$ fixing the trained network $G_{X \rightarrow Y}$ in the Stage-I. 
Here, we propose a sequential refiner $R$ that refines the image-level outputs from $G_{X \rightarrow Y}$ by rendering them to be connected more naturally in the temporal dimension.

The sequential refiner $R$ receives multiple inputs to extract correlation between consecutive frames from generated sample and source frames.
Similar to ~\cite{wang2018video}, we assume that an image sequence is generated by
Markov process that factorizes the conditional distribution which can be formalized as,
\vspace{-0.4cm}
\begin{equation}
    p(\hat{\boldsymbol{y}}|\boldsymbol{x}) = \prod_{t=1}^T p(\hat{y}_t|\{x_i\}_{i=t-L}^t, \{\hat{y}_i\}_{i=t-L}^{t-1}, G_{X\rightarrow Y}(x_t))
\end{equation}
which means that our model generates $t$-th stylized final frame $\hat{y}_t$ with consecutive $L$ prior source input frames, $L-1$ refined outputs ahead, and the intermediate output of the current frame. We set $L=2$ for our experiments. 
With the power of $R$, we design the training objective for refiner $R$ to enable the network to revise the observed defects in the intermediate frame $G_{X \rightarrow Y}(x_t)$ and consider the temporal dimension. 

Firstly, we attempt to revise empirical defects in the frame-by-frame inference of $G_{X\rightarrow Y}$. 
We observe that blurring occurs at the face boundary and the model often generates distorted background. 
We believe that this phenomenon is caused by StyleGAN as a similar issue is reported in previous works. 
This limitation does not possess a huge hindrance when it comes to image-level translation, but in the video domain, it is easy to see blurry areas around the boundary. 
We alleviate the problem by introducing a flow-based loss that applies estimated optical flow and parsing map jointly without simply separating object and background at the image level with the goal of naturally connecting frames while detecting background defects.
The objective of our suggested warp loss $\mathcal{L}_{warp}$ is as follows:
\vspace{-0.4cm}
\begin{equation}
    \mathcal{L}_{warp} = ||(M \odot Warp(x_{t-1}, f_t) + (1-M) \odot G_{X \rightarrow Y}(x_t), \hat{y}_t)||_2
\end{equation}
where $f_t$ is estimated flow from $x_{t-1}$ and $x_t$ and $M$ is predicted continuous probability parsing map for background area from $x_{t-1}$. 
This loss effectively addresses the distortion problem and does not degrade the consistency across the generated frames. We use flow estimation~\cite{ilg2017flownet} and parsing networks~\cite{yu2021bisenet} only in the training phase, thus the size and dependence of the network are not affected by auxiliary networks which differs from previous flow-based works. 

Besides, we use the additional temporal loss to suppress temporal flickering artifacts.  
As we enable our framework to leverage temporal information from consecutive input frames, we can explicitly facilitate temporal connectivity in the generated video. 
We use contextual temporal regularization loss $\mathcal{L}_{temp}$ to train the $R$ by calculating L1 distance between the outputs of $\hat{y}_{t-1}$ and $\hat{y}_t$ passed through pretrained VGG network~\cite{simonyan2014very}.

The final objective to train the sequential refiner $R$ is :
\begin{equation}
\min_{R} \lambda_{warp}\mathcal{L}_{warp} + \lambda_{temp}\mathcal{L}_{temp}.
\end{equation}

\section{Experiments}

We have experimented the model with various style datasets which consists of around a few hundred numbers of images. 
For the source video dataset, we use famous talking head video datasets VoxCeleb2~\cite{chung2018voxceleb2}. 
Please refer to the Appendix for further details about the dataset.

\begin{table}[t]
\centering
    \resizebox{.5\textwidth}{!}{
        \begin{tabular}{l | cccc |c}
        \toprule
        Metrics & 
        AgileGAN & 
        Toonify  &  
        U-GAT-IT & 
        CycleGAN  & 
        Ours \\ 
        \midrule
        CSIM ($\uparrow$) & 0.36 & 0.33 & 0.31 & 0.21 & \textbf{0.49} \\
        Eye Gaze ($\downarrow$) & 16.57 & 16.21 & 25.85 & 21.75 & \textbf{13.21} \\
        \bottomrule
        \end{tabular}}
    \vspace{-0.3cm}
\caption{Metric evaluation results on ours and baselines. }
\label{tab:metrics}
\end{table}

\noindent\textbf{Baselines.}
We compare our results with those from the following related works. 
CycleGAN~\cite{zhu2017unpaired} and U-GAT-IT~\cite{kim2019u} are image-to-image translation-based methods, and Toonify~\cite{pinkney2020resolution} and
AgileGAN~\cite{song2021agilegan} are StyleGAN-based portrait stylization approaches. Image-to-image translation frameworks are trained with VoxCeleb2 and style images dataset. The results from Toonify and AgileGAN are originally generated in $1024 \times 1024$ resolution, thus we downsample them into $256 \times 256$ to match the resolution.

\noindent\textbf{Qualitative Results.}
Fig.~\ref{fig:comparison} shows qualitative results from our model and the comparison methods. 
We provide generated samples with adjacent frames in video clips. 
As they are directly connected frames with an interval of 0.05 seconds, the resulting outputs should be connected smoothly without abrupt change.
However, we can see the undesired transform which is not only consistent with the source image but also induces unstably connected video frames from the AgileGAN and Toonify.
As StyleGAN-based approaches need a mandatory face alignment process, it further causes the incoherent output frame in video generation.  
Also, we can observe the loss of identity or incorrect facial attributes transfer.
U-GAT-IT and CycleGAN produce more stable resulting videos, but they tend to create overfitted stylized images due to the deficiency of the style dataset. 
We find that our method produces stylized videos that are more context-consistent compared to other baselines. 

\noindent\textbf{Quantitative Results.}
We conduct a quantitative evaluation with several metrics, and provide the results in Table~\ref{tab:metrics}. 
We randomly select 20 videos from the test set of VoxCeleb2 and each clip contains at least 100 frames. CSIM measures structural similarity between the source and target frames, thus higher CSIM indicates better identity preservation ability of the model. 
Note that CSIM and eye gaze distance are calculated with aligned frames only which means that we match the constraint on the StyleGAN-based model.
As evident in the table, our work outperforms all baselines on CSIM and eye gaze distance. It shows that the proposed model can create a stylized video while maintaining the source identity and well transferring the input video. 

Moreover, a user study is conducted to measure the user preference over the sample videos from comparison methods and ours. 
Similar to ~\cite{meng2021sdedit}, we ask the users to perform pairwise comparisons against our method to quantitatively evaluate the following aspects. 
The questions are 1) Stylization + Identity Preservation: which video was better stylized while maintaining the human identity of the input video? 2) Temporal Consistency: which video better maintains the temporal consistency of the input video? 
Given the source and generated frames from ours and a baseline method in a random order, we ask users to choose the better results between the two based on the provided question.
Table~\ref{tab:user} reports the user score based on 200 answers obtained from 11 users, which shows that they mostly favor our method over the baselines.
This further confirms that our method is able to generate more temporally coherent stylized video than other baselines. 

\noindent\textbf{Inference speed and Model size.} Our model has the latency of approximately 0.011 seconds for each frame, which can be considered a real-time inference speed. 
The model consists of 5.56M parameters while other StyleGAN-based frameworks have at least 30M parameters for the StyleGAN generator excluding the additional encoder for image-to-image translation.

\begin{table}[t]
\centering
    \resizebox{.5\textwidth}{!}{
        \begin{tabular}{l | cccc}
        \toprule
        Criteria & 
        AgileGAN & 
        Toonify  &  
        U-GAT-IT & 
        CycleGAN  \\
        \midrule
        Style+ID ($\uparrow$) & 67.86\% & 61.54\% & 96.43\% & 100.00\%  \\
        Temp ($\uparrow$) & 85.71\% & 76.92\% & 96.43\% & 100.00\% \\
        \bottomrule
        \end{tabular}}
    \vspace{-0.2cm}
\caption{User study results on ours and baselines. Numbers in the table indicate the percentage of users that prefer our work in the user study.}
\label{tab:user}
\end{table}

\section{Conclusion}
This work introduces a novel video portrait style transfer framework that can generate stylized videos from input frames while preserving the identity and successfully transferring the source attributes to a video of a real person. 
To effectively tackle the video stylization task, we decompose our goal into two sub-problems, portrait stylization and video generation. 
First, we train the domain mapping network by indirect utilization of the representation ability of StyleGAN and for image-level stylization. 
Then, we employ a carefully designed sequential refiner which receives multiple input images to revise image-level intermediate output and suppress the noise of the time dimension. 
The experimental results demonstrate that our network does generate plausible transferred results with fast inference time while saving computational resources significantly.

{\small
\bibliographystyle{ieee_fullname}
\bibliography{main}
}
\end{document}